\documentclass[10pt,twocolumn,letterpaper]{article}

\usepackage{iccv}
\usepackage{times}
\usepackage{epsfig}
\usepackage{graphicx}
\usepackage{amsmath}
\usepackage{amssymb}
\usepackage{booktabs}

\usepackage{multirow}
\usepackage{adjustbox}
\usepackage{xcolor}    
\usepackage{tabulary,overpic}
\usepackage{colortbl}
\usepackage{makecell}
\usepackage{subfig}
\usepackage{float}
\usepackage{url}
\usepackage{balance}
\usepackage[utf8]{inputenc} 
\usepackage[T1]{fontenc}    

\definecolor{color3}{rgb}{0.95,0.95,0.95}

\usepackage[pagebackref=true,breaklinks=true,letterpaper=true,colorlinks,bookmarks=false]{hyperref}

\iccvfinalcopy 

\usepackage[capitalize]{cleveref}
\crefname{section}{Sec.}{Secs.}
\Crefname{section}{Section}{Sections}
\Crefname{table}{Table}{Tables}
\crefname{table}{Tab.}{Tabs.}

\def\iccvPaperID{6951} 

\ificcvfinal\fi

\begin{document}

\title{Dual Aggregation Transformer for Image Super-Resolution}

\author{
  Zheng Chen$^{1}$,\enspace Yulun Zhang$^{2}$\thanks{Corresponding authors: Yulun Zhang, yulun100@gmail.com; Linghe Kong, linghe.kong@sjtu.edu.cn},\enspace Jinjin Gu$^{3,4}$,\enspace Linghe Kong$^{1*}$,\enspace Xiaokang Yang$^{1}$,\enspace Fisher Yu$^{2}$ \\
  \textsuperscript{1}Shanghai Jiao Tong University,\enspace \textsuperscript{2}ETH Z\"{u}rich,\enspace \textsuperscript{3}The University of Sydney,\enspace \textsuperscript{4}Shanghai AI Laboratory\\
}

\maketitle
\ificcvfinal\thispagestyle{empty}\fi

\begin{abstract}

Transformer has recently gained considerable popularity in low-level vision tasks, including image super-resolution (SR). These networks utilize self-attention along different dimensions, spatial or channel, and achieve impressive performance. This inspires us to combine the two dimensions in Transformer for a more powerful representation capability. Based on the above idea, we propose a novel Transformer model, Dual Aggregation Transformer (DAT), for image SR. Our DAT aggregates features across spatial and channel dimensions, in the inter-block and intra-block dual manner. Specifically, we alternately apply spatial and channel self-attention in consecutive Transformer blocks. The alternate strategy enables DAT to capture the global context and realize inter-block feature aggregation. Furthermore, we propose the adaptive interaction module (AIM) and the spatial-gate feed-forward network (SGFN) to achieve intra-block feature aggregation. AIM complements two self-attention mechanisms from corresponding dimensions. Meanwhile, SGFN introduces additional non-linear spatial information in the feed-forward network. Extensive experiments show that our DAT surpasses current methods. Code and models are obtainable at~\url{https://github.com/zhengchen1999/DAT}.

\end{abstract}


\setlength{\abovedisplayskip}{1pt}
\setlength{\belowdisplayskip}{1pt}

\vspace{-5mm}
\section{Introduction}
Single image super-resolution (SR) is a traditional low-level vision task that focuses on recovering a high-resolution (HR) image from a low-resolution (LR) counterpart. As an ill-posed problem with multiple potential solutions for a given LR input, various approaches have emerged to tackle this challenge in recent years. Many of these methods utilize convolutional neural networks (CNNs)~\cite{dong2014learning,zhang2018image,dai2019secondSAN,mei2020imageCSNLN}. However, the convolution adopts a local mechanism, which hinders the establishment of global dependencies and restricts the performance of the model.

\begin{figure}[t]
\centering
\resizebox{0.50\textwidth}{!}{
\begin{tabular}{cc}
\hspace{-0.45cm}
\begin{adjustbox}{valign=t}
\begin{tabular}{c}
\includegraphics[width=0.226\textwidth]{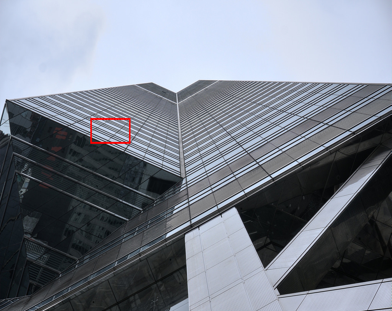}
\\
Urban100: img\_059
\end{tabular}
\end{adjustbox}
\hspace{-0.46cm}
\begin{adjustbox}{valign=t}
\begin{tabular}{cccc}
\includegraphics[width=0.149\textwidth]{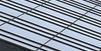} \hspace{-4mm} &
\includegraphics[width=0.149\textwidth]{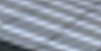} \hspace{-4mm} &
\includegraphics[width=0.149\textwidth]{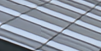} \hspace{-4mm}
\\
HR  \hspace{-4mm} &
Bicubic \hspace{-4mm} &
CSNLN~\cite{mei2020imageCSNLN} \hspace{-4mm} &
\\
\includegraphics[width=0.149\textwidth]{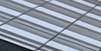} \hspace{-4mm} &
\includegraphics[width=0.149\textwidth]{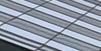} \hspace{-4mm} &
\includegraphics[width=0.149\textwidth]{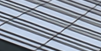} \hspace{-4mm}  
\\ 
SwinIR~\cite{liang2021swinir}  \hspace{-4mm} &
CAT-A~\cite{chen2022cross}  \hspace{-4mm} &
DAT (Ours)\hspace{-4mm}
\\
\end{tabular}
\end{adjustbox}

\end{tabular}
}
\vspace{-3mm}
\caption{\small{Visual comparison ($\times$4) on Urban100. CSNLN, SwinIR, and CAT-A suffer from blurring artifacts.}}
\label{fig:sr_vs_1}
\vspace{-6mm}
\end{figure}

Recently, Transformer proposed in natural language processing (NLP) has performed notably in multiple high-level vision tasks~\cite{dosovitskiy2020image,wang2021pyramid,liu2021swin,ding2022davit,chen2022mixformer}. The core of Transformer is the self-attention (SA) mechanism, which is capable of establishing global dependencies. This property alleviates the limitations of CNN-based algorithms. Considering the potential of Transformer, some researchers attempt to apply Transformer to low-level tasks~\cite{liang2021swinir,zamir2021restormer,wang2022uformer,chen2022cross}, including image SR. They explore efficient usages of Transformer on high-resolution images from different perspectives to mitigate the high complexity of global self-attention~\cite{dosovitskiy2020image}. For the spatial aspect, some methods~\cite{liang2021swinir,zhang2022efficient,chen2022cross} apply local spatial windows to limit the scope of self-attention. For the channel aspect, the ``transposed'' attention~\cite{zamir2021restormer} is proposed, which calculates self-attention along the channel dimension rather than the spatial dimension. These methods all exhibit remarkable results due to the strong modeling ability in their respective dimensions. Spatial window self-attention (SW-SA) is able to model fine-grained spatial relationships between pixels. Channel-wise self-attention (CW-SA) can model relationships among feature maps, thus exploiting global image information. Generally, both extracting spatial information and capturing channel context are crucial to the performance of Transformer in image SR.

Motivated by the aforementioned findings, we propose the Dual Aggregation Transformer (DAT) for image SR. 
Our DAT aggregates spatial and channel features via the inter-block and intra-block dual way to obtain powerful representation capability. Specifically, we alternately apply spatial window and channel-wise self-attention in successive dual aggregation Transformer blocks (DATBs). Through this alternate strategy, our DAT can capture both spatial and channel context and realize inter-block feature aggregation between different dimensions. Moreover, the two self-attention mechanisms complement each other. Spatial window self-attention enriches the spatial expression of each feature map, helping to model channel dependencies. Channel-wise self-attention provides the global information between features for spatial self-attention, expanding the receptive field of window attention.

Meanwhile, since self-attention mechanisms focus on modeling global information, we incorporate convolution to self-attention in parallel, to complement Transformer with the locality. To enhance the fusion of the two branches and aggregate both spatial and channel information within a single self-attention module, we propose the adaptive interaction module (AIM). It consists of two interaction operations, spatial-interaction (S-I) and channel-interaction (C-I), which act between two branches to exchange information. Through S-I and C-I, the AIM adaptively re-weight the feature maps of two branches from the spatial or channel dimension, according to different self-attention mechanisms. Besides, with AIM, we design two new self-attention mechanisms, adaptive spatial self-attention (AS-SA) and adaptive channel self-attention (AC-SA), based on the spatial window and channel-wise self-attention, respectively.

Furthermore, another component of the Transformer block, the feed-forward network (FFN)~\cite{vaswani2017attention}, extracts features through fully-connected layers. It ignores modeling spatial information. In addition, the redundant information between channels obstructs further advances in feature representation learning. To cope with these issues, we design the spatial-gate feed-forward network (SGFN), which introduces the spatial-gate (SG) module between two fully-connected layers of FFN. The SG module is a simple gating mechanism (depth-wise convolution and element-wise multiplication). The input feature of SG is partitioned into two segments along the channel dimension for convolution and multiplicative bypass. Our SG module can complement FFN with additional non-linear spatial information and relieve channel redundancy. In general, based on AIM and SGFN, DAT can realize {intra-block} feature aggregation.

Overall, with the above three designs, our DAT can aggregate spatial and channel information through the inter-block and intra-block dual way to achieve strong feature expressions. Consequently, as displayed in Fig.~\ref{fig:sr_vs_1}, our DAT achieves superior visual results against recent state-of-the-art SR methods. Our contributions are three-fold:

\begin{itemize}
\item We design a new image SR model, dual aggregation Transformer (DAT). Our DAT aggregates spatial and channel features in the inter-block and intra-block dual manner to obtain powerful representation ability.
\item We alternately adopt spatial and channel self-attention, realizing inter-block spatial and channel feature aggregation. Moreover, we propose AIM and SGFN to achieve intra-block feature aggregation.
\item We conduct extensive experiments to demonstrate that our DAT outperforms state-of-the-art methods, while retaining lower complexity and model size.
\end{itemize}

\begin{figure*}[t]
\centering
\begin{tabular}{c}
\hspace{-1.5mm}\includegraphics[width=0.99\linewidth]{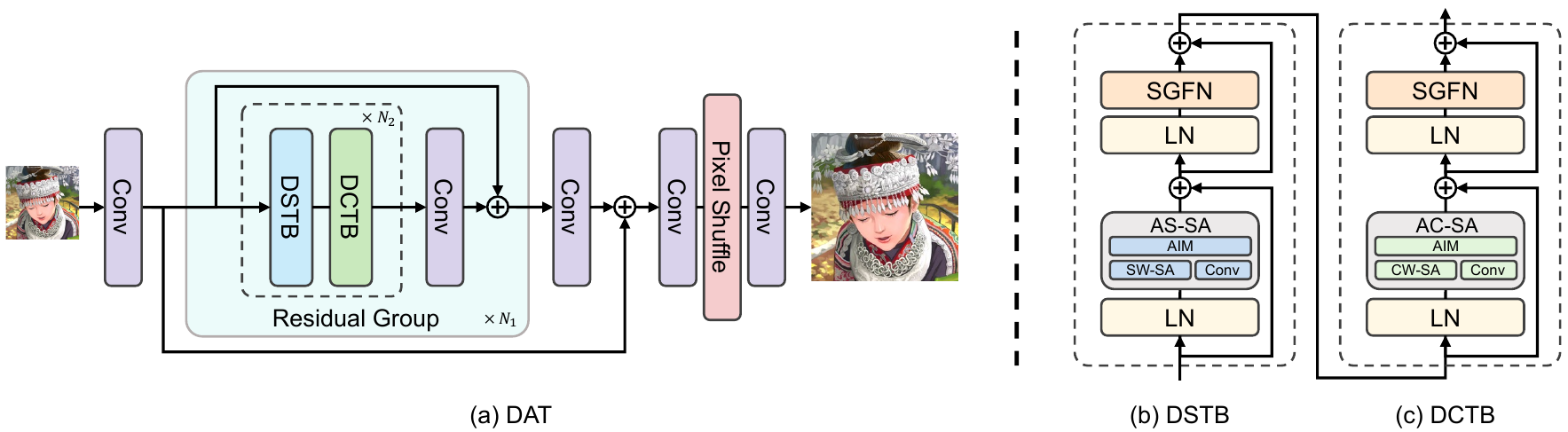} \\
\end{tabular}
\vspace{-3mm}
\caption{\small{The network architecture of our method. (a) Dual aggregation Transformer (DAT). (b) Dual spatial Transformer block (DSTB). (c) Dual channel Transformer block (DCTB). DSTB and DCTB are two consecutive dual aggregation Transformer blocks (DATBs).}}
\label{fig:architecture}
\vspace{-4mm}
\end{figure*}

\section{Related Work}
\vspace{-1.5mm}
\noindent \textbf{Image Super-Resolution.}
Deep CNN-based approaches exhibit significant efficacy in the field of image SR. SRCNN~\cite{dong2014learning} is the pioneering work, which first utilizes CNN and outperforms traditional approaches. Following this attempt, substantial dedication has been invested in deepening the layer of the network for better performance. For instance, RCAN~\cite{zhang2018image} designs residual in residual structure~\cite{he2016deep} and builds a 400+ layers model. Besides, attention mechanisms~\cite{zhang2019rnan,niu2020singleHAN,mei2020imageCSNLN,chen2021attention} in terms of spatial or channel dimensions are adopted to achieve further improvement in modeling ability. However, it is still hard for the majority of CNN-based methods to effectively model global dependencies in both spatial and channel dimensions.

\noindent \textbf{Vision Transformer.} 
Transformer demonstrates remarkable performance in high-level vision tasks~\cite{dosovitskiy2020image,wang2021pyramid,tu2022maxvit}. A series of Transformer-based methods are proposed to improve the efficiency and effectiveness of Transformer for high-level tasks. Swin Transformer~\cite{liu2021swin} applies local windows to limit the attention scope and shift operations to increase the window interaction. DaViT~\cite{ding2022davit} proposes dual self-attention to capture global context with linear complexity. Due to the remarkable performance of Transformer, researchers have been exploring the utilization of Transformer in low-level vision~\cite{wang2022uformer,chen2021preIPT,chen2023recursive,zhang2022efficient}. SwinIR~\cite{liang2021swinir} utilizes spatial window self-attention and the shift operation, following the design of Swin Transformer. Restormer~\cite{zamir2021restormer} operates self-attention along channel dimensions and applies the U-Net architecture~\cite{ronneberger2015u}. These methods remarkably outperform CNN-based methods. It reveals that both spatial and channel information are important for performance.

\noindent \textbf{Feature Aggregation.}
Several works have attempted to aggregate features among different dimensions in multiple vision tasks~\cite{woo2018cbam,zhang2019rnan,tolstikhin2021mlp} for performance improvement. In CNN, researchers apply attention mechanisms on both spatial and channel dimensions to enhance feature expressions, such as SCA-CNN~\cite{chen2017sca} and DANet~\cite{fu2019dual}. In Transformer~\cite{dosovitskiy2020image}, the spatial self-attention models long-range dependencies between pixels. Some researchers explore introducing channel attention in Transformer~\cite{zhou2022understanding,chen2022mixformer} to aggregate spatial and channel information. It effectively boosts the modeling ability of Transformer. In our work, we alternately utilize spatial and channel self-attention to achieve inter-block feature aggregation. Moreover, we propose AIM and SGFN to obtain intra-block feature aggregation.

\vspace{-3mm}
\section{Method}
\vspace{-1.5mm}
In this section, we begin by introducing the architecture of dual aggregation Transformer (DAT). Subsequently, we elaborate on the core component of DAT: Dual Aggregation Transformer Block (DATB). Finally, we analyze dual feature aggregation across spatial and channel dimensions.

\subsection{Architecture}
The overall network of the proposed DAT comprises three modules: shallow feature extraction, deep feature extraction, and image reconstruction, as illustrated in Fig.~\ref{fig:architecture}. Initially, given a low-resolution (LR) input image ${I_{LR}}$$\in$$\mathbb{R}^{H \times W \times 3}$, we employ a convolution layer to process it and generate the shallow feature ${F_S}$$\in$$\mathbb{R}^{H \times W \times C}$. Notations $H$ and $W$ denote the height and width of the input image, while $C$ represents the number of feature channels.

Subsequently, the shallow feature ${F_S}$ undergoes processing within the deep feature extraction module to acquire the deep feature ${F_{D}}$$\in$$\mathbb{R}^{H \times W \times C}$. The module is stacked by multiple residual groups (RGs), with the total number $N_1$. Meanwhile, to ensure training stability, a residual strategy is employed in the module. Each RG contains $N_2$ pairs of dual aggregation Transformer blocks (DATBs). As depicted in Fig.~\ref{fig:architecture}, each DATB pair contains two Transformer blocks, utilizing spatial and channel self-attention, respectively. A convolution layer is introduced at the end of RG to refine features extracted from Transformer blocks. Besides, for each RG, the residual connection is employed. 

Finally, we reconstruct the high-resolution (HR) output image ${I_{HR}}$$\in$$\mathbb{R}^{H_{out} \times W_{out} \times 3}$ through the reconstruction module, where $H_{out}$ is the height of the output image, and $W_{out}$ denotes image width. In this module, the deep feature $F_{D}$ is upsampled through the pixel shuffle method~\cite{shi2016real}. And convolution layers are employed to aggregate features before and after the upsampling operation.

\begin{figure*}[t]
\centering
\begin{tabular}{c}
\hspace{-1.5mm}\includegraphics[width=0.99\linewidth]{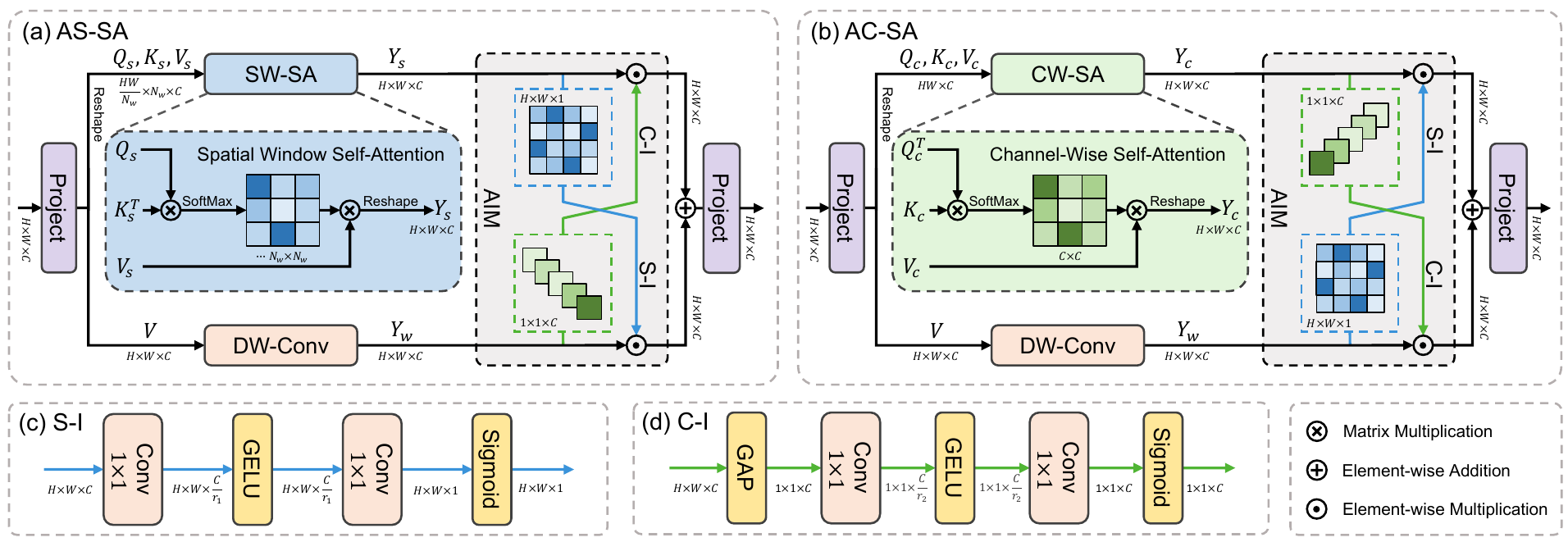} \\
\end{tabular}
\vspace{-3mm}
\caption{\small{Illustration of adaptive interaction module (AIM). (a) Adaptive spatial self-attention (AS-SA), SW-SA equipped with AIM. (b) Adaptive channel self-attention (AC-SA), CW-SA equipped with AIM. (c) Spatial-interaction (S-I). (d) Channel-interaction (C-I).}}
\label{fig:aim}
\vspace{-3mm}
\end{figure*}

\subsection{Dual Aggregation Transformer Block}
The dual aggregation Transformer block (DATB) is the core component of our proposed method. There are two kinds of DATB: dual spatial Transformer block (DSTB) and dual channel Transformer block (DCTB), as depicted in Fig.~\ref{fig:architecture}. DSTB and DCTB are based on spatial window self-attention and channel-wise self-attention, respectively. By alternately organizing DSTB and DCTB, DAT can realize inter-block feature aggregation between spatial and channel dimensions. Moreover, the adaptive interaction module (AIM) and the spatial-gate feed-forward network (SGFN) are proposed to achieve intra-block feature aggregation. Next, We describe the details below.

\noindent \textbf{Spatial Window Self-Attention.} 
The spatial window self-attention (SW-SA) computes attention within windows. As displayed in Fig.~\ref{fig:aim}\textcolor{red}{(a)}, given the input $X$$\in$$\mathbb{R}^{H \times W \times C}$, we generate $query$, $key$, and $value$ matrices (denoted as $Q$, $K$, and $V$, respectively)  through linear projection, where all matrices are in $\mathbb{R}^{H \times W \times C}$ space. The process is defined as
\vspace{1mm}
\begin{equation}
\begin{gathered}
Q = X W_Q, K = X W_K, V =X W_V,
\label{eq:qkv}
\end{gathered}
\vspace{1mm}
\end{equation}
where ${W_Q}, {W_K}, {W_V}$$\in$$\mathbb{R}^{C \times C}$ are linear projections with biases omitted. Subsequently, we partition $Q$, $K$, and $V$ into non-overlapping windows, and flat each window, which contains $N_w$ pixels. We denote the reshaped projection matrices as $Q_{s}$, $K_{s}$, and $V_s$ (all sizes are $\mathbb{R}^{\frac{HW}{N_w} \times N_w \times C}$). Then, we split them into $h$ heads: $Q_s$=$[Q_s^1,\dots,Q_s^h]$, $K_s$=$[K_s^1,\dots,K_s^h]$, and $V_s$=$[V_s^1,\dots,V_s^h]$. The dimension of each head is $d$$=$$\frac{C}{h}$. The illustration in Fig.~\ref{fig:aim}\textcolor{red}{(a)} is the situation with $h$=1, where certain details are omitted for simplicity. The output $Y_s^i$ for the $i$-th head is defined as
\vspace{1mm}
\begin{equation}
\begin{gathered}
Y_s^i = {\rm softmax}({{Q_s^i} (K_s^i)^T}/{\sqrt{d}} + D)\cdot V_s^i,
\label{eq:s-attention}
\end{gathered}
\vspace{1mm}
\end{equation}
where $D$ denotes the relative position encoding~\cite{wang2021crossformer}. Finally, we obtain the feature $Y_s$$\in$$\mathbb{R}^{H \times W \times C}$ by reshaping and concatenating all $Y_s^i$. The process is formulated as
\begin{equation}
\begin{gathered}
Y_s = {\rm concat}(Y_s^1,\dots,Y_s^h),\\
{\rm SW\mbox{-}SA}(X) = Y_s W_p,
\label{eq:sw-sa}
\end{gathered}
\end{equation}
where $W_p$$\in$$\mathbb{R}^{C \times C}$ is the linear projection to fuse all features. Moreover, following the design of Swin Transformer~\cite{liu2021swin}, we employ shift window operations by default to capture more spatial information.

\noindent \textbf{Channel-Wise Self-Attention.} 
The self-attention mechanism in the channel-wise self-attention (CW-SA) is performed along the channel dimension. Following previous works~\cite{zamir2021restormer,ali2021xcit}, we divide channels into heads and apply attention per head separately. As described in Fig.~\ref{fig:aim}\textcolor{red}{(b)}, given input $X$, we apply linear projection to generate $query$, $key$, and $value$ matrices, and reshape all of them to size $\mathbb{R}^{HW\times C}$. We denote the reshaped matrices as $Q_{c}$, $K_{c}$, and $V_c$. Same as the operation in SW-SA, we divide the projection vector into $h$ heads. Note that Fig.~\ref{fig:aim}\textcolor{red}{(b)} also depicts the case of $h$=1 for simplicity. Then the channel self-attention process of $i$-th head can be calculated as
\vspace{1mm}
\begin{equation}
\begin{gathered}
{Y_c^i} = V_c^i \cdot {\rm softmax}({({Q_c^i})^T{K_c^i}}/{\alpha}),
\label{eq:cw-sa}
\end{gathered}
\vspace{1mm}
\end{equation}
where ${Y_c^i}$$\in$$\mathbb{R}^{HW \times d}$ is the output for the $i$-th head, and $\alpha$ is a learnable temperature parameter to adjust the inner products before the softmax function. Finally, we get the attention feature $Y_c$$\in$$\mathbb{R}^{H \times W \times C}$ by concatenating and reshaping all ${Y_c^i}$. The process definition is the same as Eq.~\eqref{eq:sw-sa}.

\vspace{0.2em}
\noindent \textbf{Adaptive Interaction Module.} 
\label{sec:aim}
Since that self-attention focuses on capturing global features, we incorporate a convolution branch parallel to the self-attention module to introduce locality into Transformer. However, simply adding the convolution branch cannot effectively couple global and local features. Moreover, although alternate execution of SW-SA and CW-SA can capture both spatial and channel features, information of different dimensions still cannot be effectively utilized within a single self-attention.

To overcome these issues, we propose the adaptive interaction module (AIM), which acts between two branches, shown in Fig.~\ref{fig:aim}. It adaptively re-weights features of two branches from the spatial or channel dimension, according to the kind of self-attention mechanism. Therefore, the two branch features can be better fused. Also, both spatial and channel information can be aggregated in a single attention module. Based on AIM, we design two new self-attention mechanisms, named adaptive spatial self-attention (AS-SA) and adaptive channel self-attention (AC-SA). 

\textbf{Firstly}, we operate the parallel depth-wise convolution (DW-Conv) on $value$ of self-attention ($V$, defined in Eq.~\eqref{eq:qkv}), to establish the direct connection between self-attention and convolution. We denote the convolution output as $Y_w$$\in$$\mathbb{R}^{H\times W\times C}$.
\textbf{Then}, we introduce the AIM to adaptively adjust two features. Specifically, the AIM is based on attention mechanisms~\cite{hu2018squeeze}, including two interaction operations: spatial-interaction (S-I) and channel-interaction (C-I). Given two input features, $A$$\in$$\mathbb{R}^{H \times W \times C}$ and $B$$\in$$\mathbb{R}^{H \times W \times C}$, spatial-interaction calculates the spatial attention map (denoted as S-Map, size is $\mathbb{R}^{H \times W \times1}$) of one input (here is $B$). Channel-interaction infers the channel attention map (denoted as C-Map, size is $\mathbb{R}^{1\times 1 \times C}$). The operations are illustrated in Fig.~\ref{fig:aim}\textcolor{red}{(c, d)}, calculated as
\begin{equation}
\begin{gathered}
{\rm S\mbox{-}Map}(B) =   f(W_2 \sigma(W_1 B)), \\
{\rm C\mbox{-}Map}(B) =   f(W_4 \sigma(W_3 H_{GP}(B))), 
\label{eq:s-map&c-map}
\end{gathered}
\end{equation}
where $H_{GP}$ denotes the global average pooling, $f(\cdot)$ is the sigmoid function, and $\sigma(\cdot)$ represents the GELU function. $W_{(\cdot)}$ indicates the weight of the point-wise convolution for downscaling or upscaling channel dimensions. The reduction ratios of $W_1$ and $W_2$ are $r_1$, $\frac{C}{r_1}$, respectively. $W_3$ has a reduction ratio $r_2$, and $W_4$ has an increasing ratio $r_2$. Subsequently, the attention map is applied to another input (here is $A$), enabling the interaction. The process is formulated as
\begin{equation}
\begin{gathered}
{\rm S\mbox{-}I}(A,B) = A \odot {\rm S\mbox{-}Map}(B),\\
{\rm C\mbox{-}I}(A,B) = A \odot {\rm C\mbox{-}Map}(B),
\label{eq:adaptive-interaction}
\end{gathered}
\end{equation}
where $\odot$ denotes the element-wise multiplication. \textbf{Finally}, with AIM, we design two new self-attention mechanisms, AS-SA and AC-SA, based on SW-SA and CW-SA, respectively. As depicted in Fig.~\ref{fig:aim}\textcolor{red}{(a, b)}, for SW-SA, we introduce channel-spatial interaction between the two branches. For CW-SA, we apply spatial-channel interaction. Given the input $X$$\in$$\mathbb{R}^{H \times W \times C}$, the process is defined as
\begin{equation}
\begin{gathered}
{\rm AS\mbox{-}SA}(X) = {\rm (C\mbox{-}I}(Y_{s},Y_{w}) + {\rm S\mbox{-}I}(Y_{w},Y_{s}))W_p,\\
{\rm AC\mbox{-}SA}(X) = {\rm (S\mbox{-}I}(Y_{c},Y_{w}) + {\rm C\mbox{-}I}(Y_{w},Y_{c}))W_p,
\label{eq:aim}
\end{gathered}
\end{equation}
where $Y_s$, $Y_c$, and $Y_w$ are the outputs of SW-SA, CW-SA, and DW-Conv defined above. $W_p$ is the projection matrix the same as Eq.~\eqref{eq:sw-sa}. Besides, we collectively refer to AC-SA and AS-SA as adaptive self-attention (A-SA) for simplicity.

With AIM, our proposed AS-SA and AC-SA have two advantages over SW-SA and CW-SA. 
\emph{\textbf{Firstly, better coupling of local (convolution) and global (attention).}} Convolution aggregates locality information in the neighbourhood, while self-attention models long-range dependencies. However, considering the feature misalignment between the two branches, simple addition is not convincing enough. Through adaptive interaction, the outputs of the two branches can be adaptively adjusted to fit each other, thus achieving better feature fusion.
\emph{\textbf{Secondly, stronger modeling ability.}} For AS-SA, the complementary clues improve its channel-wise modeling ability, through channel-interaction. For AC-SA, the representation capability is boosted by additional spatial knowledge, through spatial-interaction. Furthermore, through adaptive interaction, global information can flow from self-attention to the convolution branch. It enhances the output of convolution.

\begin{figure}[t]
\centering
\begin{tabular}{c}
\hspace{-2mm}\includegraphics[width=0.99\linewidth]{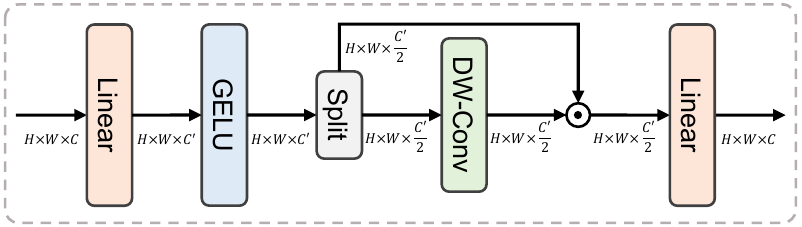} \\
\end{tabular}
\vspace{-4mm}
\caption{\small{Illustration of spatial-gate feed-forward network.}}
\label{fig:SGFN}
\vspace{-6mm}
\end{figure}

\vspace{0.2em}
\noindent \textbf{Spatial-Gate Feed-Forward Network.} 
The feed-forward network (FFN)~\cite{vaswani2017attention} has a non-linear activation and two linear projection layers to extract features. However, it ignores modeling spatial information. Besides, the redundant information in channels hinders feature expression competence. To overcome the above limitations, we propose the spatial-gate feed-forward network (SGFN), introducing spatial-gate (SG) to FFN. As shown in Fig.~\ref{fig:SGFN}, our SG module is a simple gate mechanism, consisting of depth-wise convolution and element-wise multiplication. Along the channel dimension, we divide the feature map into two parts for convolutional and multiplicative bypass. Overall, given the input $\hat{X}$$\in$$\mathbb{R}^{H \times W \times C}$, SGFN is computed as
\begin{equation}
\begin{gathered}
\hat{X'} = \sigma(W_p^1 \hat{X}),\quad [\hat{X_1'}, \hat{X_2'}] = \hat{X'},\\
{\rm SGFN}(\hat{X}) = W_p^2(\hat{X_1'} \odot (W_d \hat{X_2'})),
\label{eq:sgfn}
\end{gathered}
\end{equation}
where $W_p^1$ and $W_p^2$ represent linear projections, $\sigma$ denotess the GELU function, and $W_d$ is the learnable parameters of the depth-wise convolution. Both $\hat{X_1'}$ and $\hat{X_2'}$ are in $\mathbb{R}^{H \times W \times \frac{C'}{2}}$ space, where $C'$ denotes the hidden dimension in SGFN. Compared with FFN, our SGFN is able to capture non-linear spatial information and ease the channel redundancy of fully-connected layers. Moreover, different from previous works~\cite{liu2021pay,chen2022simple,tu2022maxim}, our SG module utilizes depth-wise convolution to maintain computational efficiency.

\vspace{0.2em}
\noindent \textbf{Dual Aggregation Transformer Block.} 
Our dual aggregation Transformer block (DATB) is equipped with the adaptive self-attention (A-SA) and the spatial-gate feed-forward network (SGFN). Given the input $X_{l-1}$$\in$$\mathbb{R}^{H \times W \times C}$ of the $l$-th block, the block is defined as
\begin{equation}
\begin{split}
& {X_{l}}' = {\rm A\mbox{-}SA}({\rm LN}(X_{l-1})) + X_{l-1}, \\
& X_{l} = {\rm SGFN}({\rm LN}({X_{l}}')) + {X_{l}}', 
\label{eq:block}
\end{split}
\end{equation}
where $X_{l}$ is the output features, and ${\rm LN}(\cdot)$ is the LayerNorm layer. Since A-SA includes AS-SA and AC-SA, there are two types of DATB, dual spatial Transformer block (DSTB) and dual channel Transformer block (DCTB). DSTB applies AS-SA, while DCTB adopts AC-SA.

\subsection{Dual Feature Aggregation}
Our DAT is capable of aggregating the spatial and channel features through the inter-block and intra-block dual manner to obtain powerful feature representations. 

\noindent \textbf{Inter-block Aggregation.} 
DAT alternately adopts DSTB and DCTB to capture features in both dimensions, and make use of their complementary advantages. Specifically, DSTB models long-range spatial context, enhancing the spatial expression of each feature map. Meanwhile, DCTB can better build channel dependencies. DCTB models global channel context, which in turn helps DSTB to capture spatial features and also enlarge the receptive field. Consequently, both spatial and channel information flow between consecutive Transformer blocks and thus can be aggregated.

\noindent \textbf{Intra-block Aggregation.} 
AIM can complement spatial window self-attention with channel knowledge, and enhance channel-wise self-attention from the spatial dimension. Moreover, SGFN is able to introduce additional non-linear spatial information into FFN that only models channel relationships. Therefore, DAT can aggregate spatial and channel features in each Transformer block.

\begin{table*}[t]
\centering
\subfloat[\small  Ablation study of alternate strategy.\label{abl:1}]{
        \scalebox{0.75}{
        \setlength{\tabcolsep}{2mm}
        \begin{tabular}{c c c c c c c}
        \toprule
        \rowcolor{color3} CW-SA & SW-SA & Params (M) & FLOPs (G) & PSNR (dB)  & SSIM \\
        \midrule
        \checkmark & & 16.38 & 274.54 & 32.80 & 0.9340\\
          & \checkmark& 16.40  & 282.76 & 33.20 & 0.9379\\
        \checkmark & \checkmark& 16.39 & 278.15 &  33.34 & 0.9388\\
        \bottomrule
\end{tabular}}}\hspace{0mm}
\subfloat[\small  Ablation study of AIM.\label{abl:2}]{
        \scalebox{0.75}{
        \setlength{\tabcolsep}{2mm}
        \begin{tabular}{c c c c c c c}
        \toprule
        \rowcolor{color3} Baseline & DW-Conv & AIM & Params (M)  & FLOPs (G) & PSNR (dB)  & SSIM \\
        \midrule
        \checkmark & & & 16.39 & 278.15 & 33.34 & 0.9388\\
        \checkmark & \checkmark& & 16.47 & 279.31 & 33.41 & 0.9392\\
        \checkmark & \checkmark& \checkmark& 16.84 & 280.61 & 33.52 & 0.9400\\
        \bottomrule
\end{tabular}}}\vspace{-3mm}\\

\subfloat[\small  Further ablation study of AIM.\label{abl:3}]{
\scalebox{0.7}{
        \setlength{\tabcolsep}{1.5mm}
        \begin{tabular}{l cccccc}
        \toprule
        \rowcolor{color3} Model & SA$\rightarrow$Conv & Conv$\rightarrow$SA & AIM\\
        \midrule
        Params (M) & 16.65 & 16.65 & 16.84 \\
        FLOPs (G) & 279.96 & 279.96 & 280.61 \\
        PSNR (dB) & 33.43 & 33.47 & 33.52\\
        SSIM & 0.9401 & 0.9397 & 0.9400\\
        \bottomrule
\end{tabular}}}\hspace{0mm}
\subfloat[\small  Ablation study of SGFN.\label{abl:4}]{
        \scalebox{0.7}{
        \setlength{\tabcolsep}{1.5mm}
        \begin{tabular}{l c c c c}
        \toprule
        \rowcolor{color3}Model & Params (M) & FLOPs (G) & PSNR (dB)  & SSIM \\
        \midrule
        FFN & 16.84 & 280.61 & 33.52 & 0.9400\\
        SGFN \small{w/o Conv} & 14.50 & 242.39& 33.44 & 0.9390 \\
        SGFN \small{w/o Split} & 17.15 & 286.55 & 33.53 & 0.9404\\
        SGFN & 14.66 & 245.36 & 33.57 & 0.9405\\
        \bottomrule
\end{tabular}}}\hspace{0mm}
\subfloat[\small  Ablation study of different blocks.\label{abl:5}]{
        \scalebox{0.7}{
        \setlength{\tabcolsep}{3mm}
        \begin{tabular}{l cccccc}
        \toprule
        \rowcolor{color3} Model & DCTB & DSTB & DAT\\
        \midrule
        Params (M) & 14.65 & 14.67 & 14.66 \\
        FLOPs (G) & 241.75 & 248.97&  245.36\\
        PSNR (dB) & 33.26 & 33.43 & 33.57\\
        SSIM & 0.9376 & 0.9391 & 0.9405\\
        \bottomrule
\end{tabular}}}
\label{table:ablation}
\vspace{-3mm}
\caption{\small Ablation studies. The models are trained on DIV2K and Flickr2K, and tested on Urban100 ($\times$2)}
\vspace{-4mm}
\end{table*}

\vspace{-1mm}
\section{Experiments}
\vspace{-1mm}
\subsection{Experimental Settings}
\vspace{-1mm}
\noindent \textbf{Implementation Details.} 
We build two variants of DAT with different complexity, called DAT-S and DAT. For DAT-S, there are 6 residual groups (RGs), and each RG contains 3 pairs of dual aggregation Transformer blocks (DATBs) (3 DSTBs and 3 DCTBs). The attention head number, channel dimension, and channel expansion factor in SGFN are set as 6, 180, and 2 for both DSTB and DCTB. For all DSTBs, we set the window size as 8$\times$16. For DAT, we enlarge the channel expansion factor to 4 and the window size to 8$\times$32. Other settings remain the same as DAT-S.

\noindent \textbf{Data and Evaluation.} 
We follow most previous works~\cite{haris2018deep,liang2021swinir} to train and test our models. Specifically, we apply two datasets: DIV2K~\cite{timofte2017ntire} and Flickr2K~\cite{lim2017enhanced}, for training, and five benchmark datasets: Set5~\cite{bevilacqua2012low}, Set14~\cite{zeyde2012single}, B100~\cite{martin2001database}, Urban100~\cite{huang2015single}, and Manga109~\cite{matsui2017sketch}, for testing. We carry out experiments under upscaling factors: $\times$2, $\times$3, and $\times$4. LR images are generated from HR images by bicubic degradation. The evaluation of SR results is performed using two metrics: PSNR and SSIM~\cite{wang2004image}, which are calculated on the Y channel (\ie, luminance) of the YCbCr space.

\begin{figure}[t]
\scriptsize
\centering
\vspace{1.5mm}
\begin{tabular}{ccc}
\hspace{-0.4cm}
\begin{adjustbox}{valign=t}
\begin{tabular}{c}
\end{tabular}
\end{adjustbox}
\begin{adjustbox}{valign=t}
\begin{tabular}{cccccc}
\includegraphics[width=0.090\textwidth]{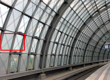} \hspace{-4mm} &
\includegraphics[width=0.090\textwidth]{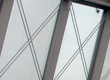} \hspace{-4mm} &
\includegraphics[width=0.090\textwidth]{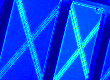} \hspace{-4mm} &
\includegraphics[width=0.090\textwidth]{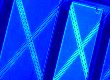} \hspace{-4mm} &
\includegraphics[width=0.090\textwidth]{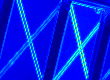} \hspace{-4mm} &
\\
\end{tabular}
\end{adjustbox}
\vspace{0.5mm}
\\
\hspace{-0.4cm}
\begin{adjustbox}{valign=t}
\begin{tabular}{c}
\end{tabular}
\end{adjustbox}
\begin{adjustbox}{valign=t}
\begin{tabular}{cccccc}
\includegraphics[width=0.090\textwidth]{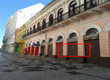} \hspace{-4mm} &
\includegraphics[width=0.090\textwidth]{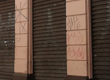} \hspace{-4mm} &
\includegraphics[width=0.090\textwidth]{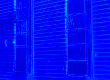} \hspace{-4mm} &
\includegraphics[width=0.090\textwidth]{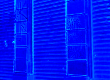} \hspace{-4mm} &
\includegraphics[width=0.090\textwidth]{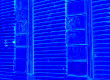} \hspace{-4mm} &
\\
HR \hspace{-4mm} &
Zoom-in \hspace{-4mm} &
CW-SA  \hspace{-4mm} &
SW-SA \hspace{-4mm} &
CW/SW-SA \hspace{-4mm} &
\\
\end{tabular}
\end{adjustbox}
\vspace{-1mm}
\\
\end{tabular}
\vspace{-2mm}
\caption{\small Visualization of different attention strategies. }
\vspace{-5mm}
\label{fig:ablation}
\end{figure}

\noindent \textbf{Training Settings.} 
We train models with patch size 64$\times$64 and batch size 32. The training iterations are 500K. We optimize models by minimizing the $L_1$ loss through Adam optimizer~\cite{kingma2014adam} ($\beta_1$=$0.9$ and ${\beta}_2$=$0.99$). We initially set the learning rate as 2$\times$10$^{-4}$, and half it at milestones: [250K,400K,450K,475K]. Furthermore, during training, we randomly utilize rotation of $90^\circ$, $180^\circ$, and $270^\circ$ and horizontal flips to augment the data. Our model is implemented based on PyTorch~\cite{paszke2017automatic} with 4 A100 GPUs.

\begin{table*}[t]
\scriptsize
\begin{center}
\begin{tabular}{l|c|cccccccccccccccc} 
\toprule[0.15em]
	\rowcolor{color3}  &  &  \multicolumn{2}{c}{Set5} & \multicolumn{2}{c}{Set14} & \multicolumn{2}{c}{B100} & \multicolumn{2}{c}{Urban100} & \multicolumn{2}{c}{Manga109}\\

	\rowcolor{color3} \multirow{-2}{*}{Method}& \multirow{-2}{*}{Scale} & PSNR & SSIM & PSNR & SSIM & PSNR & SSIM & PSNR & SSIM & PSNR & SSIM
\\
\midrule[0.15em]
EDSR~\cite{lim2017enhanced} & $\times$2 & 
38.11 & 0.9602 & 33.92 & 0.9195 & 32.32 & 0.9013 & 32.93 & 0.9351 & 39.10 & 0.9773\\
RCAN~\cite{zhang2018image} & $\times$2 &
38.27 & 0.9614 & 34.12 & 0.9216 & 32.41 & 0.9027 & 33.34 & 0.9384 & 39.44 & 0.9786\\
SAN~\cite{dai2019secondSAN} & $\times$2 &
38.31 & 0.9620 & 34.07 & 0.9213 & 32.42 & 0.9028 & 33.10 & 0.9370 & 39.32 & 0.9792\\
RFANet~\cite{liu2020residual} & $\times$2 &
38.26  & 0.9615 & 34.16 & 0.9220 & 32.41 & 0.9026 & 33.33 & 0.9389 & 39.44 & 0.9783\\
HAN~\cite{niu2020singleHAN} & $\times$2 &
38.27 & 0.9614 & 34.16 & 0.9217 & 32.41 & 0.9027 & 33.35 & 0.9385 & 39.46 & 0.9785\\
CSNLN~\cite{mei2020imageCSNLN} & $\times$2 &
38.28 & 0.9616 & 34.12 & 0.9223 & 32.40 & 0.9024 & 33.25 & 0.9386 & 39.37 & 0.9785\\
NLSA~\cite{mei2021imageNLSA} & $\times$2 &
38.34 & 0.9618 & 34.08 & 0.9231 & 32.43 & 0.9027 & 33.42 & 0.9394 & 39.59 & 0.9789\\
ELAN~\cite{zhang2022efficient} & $\times$2 &
38.36 & 0.9620 & 34.20 & 0.9228 & 32.45 & 0.9030 & 33.44 & 0.9391 & 39.62 & 0.9793\\
DFSA~\cite{magid2021dynamic} & $\times$2 &
38.38 & 0.9620 & 34.33 & 0.9232 & 32.50 & 0.9036 & 33.66 & 0.9412 & 39.98 & 0.9798\\
SwinIR~\cite{liang2021swinir} & $\times$2 &  
{38.42} & {0.9623} & {34.46} & {0.9250} &{32.53} &{0.9041} & {33.81} &{0.9427} & {39.92} & {0.9797}\\
CAT-A~\cite{chen2022cross} & $\times$2 &  
\textcolor{black}{38.51} & \textcolor{black}{0.9626} & \textcolor{black}{34.78} & \textcolor{black}{0.9265} & \textcolor{black}{32.59} & \textcolor{black}{0.9047} & \textcolor{black}{34.26} & \textcolor{black}{0.9440} & \textcolor{black}{40.10} & \textcolor{black}{0.9805}
\\ 
DAT-S (ours) & $\times$2 &  
\textcolor{black}{38.54} & \textcolor{black}{0.9627} & \textcolor{black}{34.60} & \textcolor{black}{0.9258} & \textcolor{black}{32.57} & \textcolor{black}{0.9047} & \textcolor{black}{34.12} & \textcolor{black}{0.9444} & \textcolor{black}{40.17} & \textcolor{black}{0.9804}
\\ 
DAT (ours) & $\times$2 &  
\textcolor{blue}{38.58} & \textcolor{blue}{0.9629} & \textcolor{blue}{34.81} & \textcolor{blue}{0.9272} & \textcolor{blue}{32.61} & \textcolor{blue}{0.9051} & \textcolor{blue}{34.37} & \textcolor{blue}{0.9458} & \textcolor{blue}{40.33} & \textcolor{blue}{0.9807}
\\ 
DAT+ (ours) & $\times$2 &  
\textcolor{red}{38.63} & \textcolor{red}{0.9631} & \textcolor{red}{34.86} & \textcolor{red}{0.9274} & \textcolor{red}{32.63} & \textcolor{red}{0.9053} & \textcolor{red}{34.47} & \textcolor{red}{0.9465} & \textcolor{red}{40.43} & \textcolor{red}{0.9809}
\\

\midrule
EDSR~\cite{lim2017enhanced} & $\times$3 & 
34.65 & 0.9280 & 30.52 & 0.8462 & 29.25 & 0.8093 & 28.80 & 0.8653 & 34.17 & 0.9476\\
RCAN~\cite{zhang2018image} & $\times$3 &
34.74 & 0.9299 & 30.65 & 0.8482 & 29.32 & 0.8111 & 29.09 & 0.8702 & 34.44 & 0.9499\\
SAN~\cite{dai2019secondSAN} & $\times$3 &
34.75 & 0.9300 & 30.59 & 0.8476 & 29.33 & 0.8112 & 28.93 & 0.8671 & 34.30 & 0.9494\\
RFANet~\cite{liu2020residual} & $\times$3 &
34.79 & 0.9300 & 30.67 & 0.8487 & 29.34 & 0.8115 & 29.15 & 0.8720 & 34.59 & 0.9506\\
HAN~\cite{niu2020singleHAN} & $\times$3 &
34.75 & 0.9299 & 30.67 & 0.8483 & 29.32 & 0.8110 & 29.10 & 0.8705 & 34.48 & 0.9500\\
CSNLN~\cite{mei2020imageCSNLN} & $\times$3 &
34.74 & 0.9300 & 30.66 & 0.8482 & 29.33 & 0.8105 & 29.13 & 0.8712 & 34.45 & 0.9502\\
NLSA~\cite{mei2021imageNLSA} & $\times$3 &
34.85 & 0.9306 & 30.70 & 0.8485 & 29.34 & 0.8117 & 29.25 & 0.8726 & 34.57 & 0.9508\\
ELAN~\cite{zhang2022efficient} & $\times$3 &
34.90 & 0.9313 & 30.80 & 0.8504 & 29.38 & 0.8124 & 29.32 & 0.8745 & 34.73 & 0.9517\\
DFSA~\cite{magid2021dynamic} & $\times$3 &
34.92 & 0.9312 & 30.83 & 0.8507 & 29.42 & 0.8128 & 29.44 & 0.8761 & 35.07 & 0.9525\\
SwinIR~\cite{liang2021swinir} & $\times$3 &  
{34.97} & {0.9318} & {30.93} & {0.8534} & {29.46} & {0.8145} & {29.75} & {0.8826} & {35.12} & {0.9537}\\
CAT-A~\cite{chen2022cross} & $\times$3 &  
\textcolor{black}{35.06} & \textcolor{black}{0.9326} & \textcolor{black}{31.04} & \textcolor{black}{0.8538} & \textcolor{black}{29.52} & \textcolor{black}{0.8160} & \textcolor{black}{30.12} & \textcolor{black}{0.8862} & \textcolor{black}{35.38} & \textcolor{black}{0.9546}
\\
DAT-S (ours) & $\times$3 &  
\textcolor{black}{35.12} & \textcolor{black}{0.9327} & \textcolor{black}{31.04} & \textcolor{black}{0.8543} & \textcolor{black}{29.51} & \textcolor{black}{0.8157} & \textcolor{black}{29.98} & \textcolor{black}{0.8846} & \textcolor{black}{35.41} & \textcolor{black}{0.9546}
\\
DAT (ours) & $\times$3 &  
\textcolor{blue}{35.16} & \textcolor{blue}{0.9331} & \textcolor{blue}{31.11} & \textcolor{blue}{0.8550} & \textcolor{blue}{29.55} & \textcolor{blue}{0.8169} & \textcolor{blue}{30.18} & \textcolor{blue}{0.8886} & \textcolor{blue}{35.59} & \textcolor{blue}{0.9554}
\\
DAT+ (ours) & $\times$3 &  
\textcolor{red}{35.19} & \textcolor{red}{0.9334} & \textcolor{red}{31.17} & \textcolor{red}{0.8558} & \textcolor{red}{29.58} & \textcolor{red}{0.8173} & \textcolor{red}{30.30} & \textcolor{red}{0.8902} & \textcolor{red}{35.72} & \textcolor{red}{0.9559}
\\

\midrule
EDSR~\cite{lim2017enhanced} & $\times$4 & 
32.46 & 0.8968 & 28.80 & 0.7876 & 27.71 & 0.7420 & 26.64 & 0.8033 & 31.02 & 0.9148\\
RCAN~\cite{zhang2018image} & $\times$4 &
32.63 & 0.9002 & 28.87 & 0.7889 & 27.77 & 0.7436 & 26.82 & 0.8087 & 31.22 & 0.9173\\
SAN~\cite{dai2019secondSAN} & $\times$4 &
32.64 & 0.9003 & 28.92 & 0.7888 & 27.78 & 0.7436 & 26.79 & 0.8068 & 31.18 & 0.9169\\
RFANet~\cite{liu2020residual} & $\times$4 & 
32.66 & 0.9004 & 28.88 & 0.7894 & 27.79 & 0.7442 & 26.92 & 0.8112 & 31.41 & 0.918\\
HAN~\cite{niu2020singleHAN} & $\times$4 &
32.64 & 0.9002 & 28.90 & 0.7890 & 27.80 & 0.7442 & 26.85 & 0.8094 & 31.42 & 0.9177\\
CSNLN~\cite{mei2020imageCSNLN} & $\times$4 &
32.68 & 0.9004 & 28.95 & 0.7888 & 27.80 & 0.7439 & 27.22 & 0.8168 & 31.43 & 0.9201\\
NLSA~\cite{mei2021imageNLSA} & $\times$4 &
32.59 & 0.9000 & 28.87 & 0.7891 & 27.78 & 0.7444 & 26.96 & 0.8109 & 31.27 & 0.9184\\
ELAN~\cite{zhang2022efficient} & $\times$4 &
32.75 & 0.9022 & 28.96 & 0.7914 & 27.83 & 0.7459 & 27.13 & 0.8167 & 31.68 & 0.9226\\
DFSA~\cite{magid2021dynamic} & $\times$4 &
32.79 & 0.9019 & 29.06 & 0.7922 & 27.87 & 0.7458 & 27.17 & 0.8163 & 31.88 & 0.9266\\
SwinIR~\cite{liang2021swinir} & $\times$4 & 
{32.92} & {0.9044} & {29.09} & {0.7950} & {27.92} & {0.7489} & {27.45} & {0.8254} & {32.03} & {0.9260}\\
CAT-A~\cite{chen2022cross} & $\times$4 & 
\textcolor{black}{33.08} & \textcolor{black}{0.9052} & \textcolor{black}{29.18} & \textcolor{black}{0.7960} & \textcolor{black}{27.99} & \textcolor{black}{0.7510} & \textcolor{blue}{27.89} & \textcolor{black}{0.8339} & \textcolor{black}{32.39} & \textcolor{black}{0.9285}
\\
DAT-S (ours) & $\times$4 &  
\textcolor{black}{33.00} & \textcolor{black}{0.9047} & \textcolor{black}{29.20} & \textcolor{black}{0.7962} & \textcolor{black}{27.97} & \textcolor{black}{0.7502} & \textcolor{black}{27.68} & \textcolor{black}{0.8300} & \textcolor{black}{32.33} & \textcolor{black}{0.9278}
\\
DAT (ours) & $\times$4 & 
\textcolor{blue}{33.08} & \textcolor{blue}{0.9055} & \textcolor{blue}{29.23} & \textcolor{blue}{0.7973} & \textcolor{blue}{28.00} & \textcolor{blue}{0.7515} & \textcolor{black}{27.87} & \textcolor{blue}{0.8343} & \textcolor{blue}{32.51} & \textcolor{blue}{0.9291}
\\
DAT+ (ours) & $\times$4 &  
\textcolor{red}{33.15} & \textcolor{red}{0.9062} & \textcolor{red}{29.29} & \textcolor{red}{0.7983} & \textcolor{red}{28.03} & \textcolor{red}{0.7518} & \textcolor{red}{27.99} & \textcolor{red}{0.8365} & \textcolor{red}{32.67} & \textcolor{red}{0.9301}
\\
\bottomrule[0.15em]
\end{tabular}
\vspace{-1mm}
\caption{\small{Quantitative comparison with state-of-the-art methods. The best and second-best results are coloured \textcolor{red}{red} and \textcolor{blue}{blue}.}}
\label{table:psnr_ssim_SR_5sets}
\end{center}
\vspace{-8mm}
\end{table*}

\begin{figure*}[ht]
\scriptsize
\centering
\scalebox{0.92}{
\begin{tabular}{cccc}
\hspace{-0.42cm}
\begin{adjustbox}{valign=t}
\begin{tabular}{c}
\includegraphics[width=0.216\textwidth]{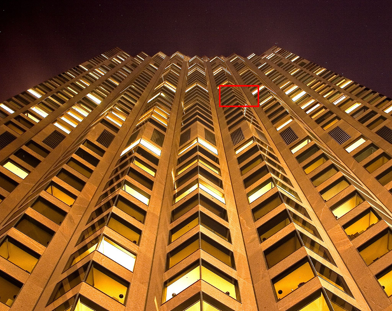}
\\
Urban100: img\_015 ($\times$4)
\end{tabular}
\end{adjustbox}
\hspace{-0.46cm}
\begin{adjustbox}{valign=t}
\begin{tabular}{cccccc}
\includegraphics[width=0.149\textwidth]{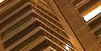} \hspace{-4mm} &
\includegraphics[width=0.149\textwidth]{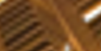} \hspace{-4mm} &
\includegraphics[width=0.149\textwidth]{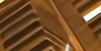} \hspace{-4mm} &
\includegraphics[width=0.149\textwidth]{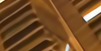} \hspace{-4mm} &
\includegraphics[width=0.149\textwidth]{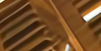} \hspace{-4mm} 
\\
HR \hspace{-4mm} &
Bicubic \hspace{-4mm} &
RCAN~\cite{zhang2018image} \hspace{-4mm} &
SAN~\cite{dai2019secondSAN} \hspace{-4mm} &
RFANet~\cite{liu2020residual} \hspace{-4mm} 
\\
\includegraphics[width=0.149\textwidth]{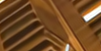} \hspace{-4mm} &
\includegraphics[width=0.149\textwidth]{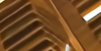} \hspace{-4mm} &
\includegraphics[width=0.149\textwidth]{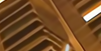} \hspace{-4mm} &
\includegraphics[width=0.149\textwidth]{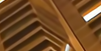} \hspace{-4mm} &
\includegraphics[width=0.149\textwidth]{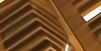} \hspace{-4mm}  
\\ 
HAN~\cite{niu2020singleHAN} \hspace{-4mm} &
CSNLN~\cite{mei2020imageCSNLN} \hspace{-4mm} &
SwinIR~\cite{liang2021swinir}  \hspace{-4mm} &
CAT-A~\cite{chen2022cross}  \hspace{-4mm} &
DAT (ours) \hspace{-4mm}
\\
\end{tabular}
\end{adjustbox}
\\
\hspace{-0.42cm}
\begin{adjustbox}{valign=t}
\begin{tabular}{c}
\includegraphics[width=0.216\textwidth]{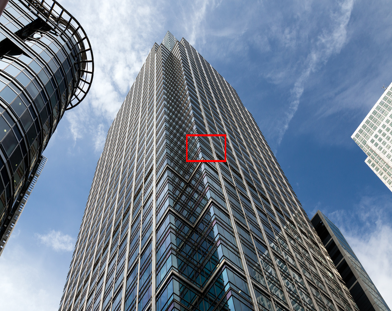}
\\
Urban100: img\_047 ($\times$4)
\end{tabular}
\end{adjustbox}
\hspace{-0.46cm}
\begin{adjustbox}{valign=t}
\begin{tabular}{cccccc}
\includegraphics[width=0.149\textwidth]{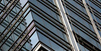} \hspace{-4mm} &
\includegraphics[width=0.149\textwidth]{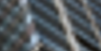} \hspace{-4mm} &
\includegraphics[width=0.149\textwidth]{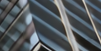} \hspace{-4mm} &
\includegraphics[width=0.149\textwidth]{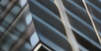} \hspace{-4mm} &
\includegraphics[width=0.149\textwidth]{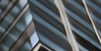} \hspace{-4mm} 
\\
HR \hspace{-4mm} &
Bicubic \hspace{-4mm} &
RCAN~\cite{zhang2018image} \hspace{-4mm} &
SAN~\cite{dai2019secondSAN} \hspace{-4mm} &
RFANet~\cite{liu2020residual} \hspace{-4mm} 
\\
\includegraphics[width=0.149\textwidth]{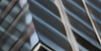} \hspace{-4mm} &
\includegraphics[width=0.149\textwidth]{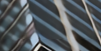} \hspace{-4mm} &
\includegraphics[width=0.149\textwidth]{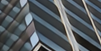} \hspace{-4mm} &
\includegraphics[width=0.149\textwidth]{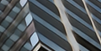} \hspace{-4mm} &
\includegraphics[width=0.149\textwidth]{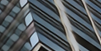} \hspace{-4mm}  
\\ 
HAN~\cite{niu2020singleHAN} \hspace{-4mm} &
CSNLN~\cite{mei2020imageCSNLN} \hspace{-4mm} &
SwinIR~\cite{liang2021swinir}  \hspace{-4mm} &
CAT-A~\cite{chen2022cross}  \hspace{-4mm} &
DAT (ours) \hspace{-4mm}
\\
\end{tabular}
\end{adjustbox}
\\
\hspace{-0.42cm}
\begin{adjustbox}{valign=t}
\begin{tabular}{c}
\includegraphics[width=0.216\textwidth]{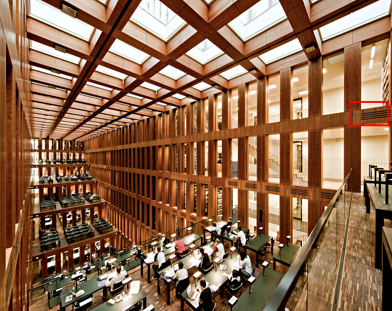}
\\
Urban100: img\_049 ($\times$4)
\end{tabular}
\end{adjustbox}
\hspace{-0.46cm}
\begin{adjustbox}{valign=t}
\begin{tabular}{cccccc}
\includegraphics[width=0.149\textwidth]{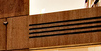} \hspace{-4mm} &
\includegraphics[width=0.149\textwidth]{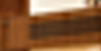} \hspace{-4mm} &
\includegraphics[width=0.149\textwidth]{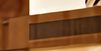} \hspace{-4mm} &
\includegraphics[width=0.149\textwidth]{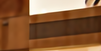} \hspace{-4mm} &
\includegraphics[width=0.149\textwidth]{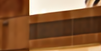} \hspace{-4mm} 
\\
HR \hspace{-4mm} &
Bicubic \hspace{-4mm} &
RCAN~\cite{zhang2018image} \hspace{-4mm} &
SAN~\cite{dai2019secondSAN} \hspace{-4mm} &
RFANet~\cite{liu2020residual} \hspace{-4mm} 
\\
\includegraphics[width=0.149\textwidth]{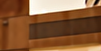} \hspace{-4mm} &
\includegraphics[width=0.149\textwidth]{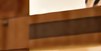} \hspace{-4mm} &
\includegraphics[width=0.149\textwidth]{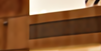} \hspace{-4mm} &
\includegraphics[width=0.149\textwidth]{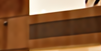} \hspace{-4mm} &
\includegraphics[width=0.149\textwidth]{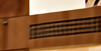} \hspace{-4mm}  
\\ 
HAN~\cite{niu2020singleHAN} \hspace{-4mm} &
CSNLN~\cite{mei2020imageCSNLN} \hspace{-4mm} &
SwinIR~\cite{liang2021swinir}  \hspace{-4mm} &
CAT-A~\cite{chen2022cross}  \hspace{-4mm} &
DAT (ours) \hspace{-4mm}
\\
\end{tabular}
\end{adjustbox}

\end{tabular} }
\vspace{-3mm}
\caption{\small{Visual comparison for image SR ($\times$4) in some challenging cases.}}
\label{fig:sr_vs_2}
\vspace{-6mm}
\end{figure*}

\subsection{Ablation Study}
We train models on the dataset DIV2K~\cite{timofte2017ntire} and Flickr2K~\cite{lim2017enhanced} and test them on Urban100~\cite{huang2015single} in the ablation study. For a fair comparison, all models have the same implementation details (\eg, residual group number) as DAT. The iterations are 300K. Besides, we set the output size as 3$\times$256$\times$256 to compute FLOPs.

\noindent \textbf{Alternate Strategy.} 
To investigate the effect of the strategy for alternating using SW-SA and CW-SA, we carry out several experiments, and list results in Table~\ref{abl:1}.
The first and second rows of the table mean we replace all attention modules in DAT with CW-SA or SW-SA, where SW-SA adopts the 8$\times$8 window size. The third row represents alternately applying two SA in consecutive Transformer blocks in DAT. Moreover, all models apply the regular FFN~\cite{vaswani2017attention} and do not adopt AIM in SA. Comparing the three models, we can observe that the model utilizing SW-SA outperforms the model using CW-SA. Furthermore, alternately applying two SA can get the best performance of 33.34 dB. It indicates that exploiting both channel and spatial information is crucial to accurate image restoration. 

Additionally, we visualize the last feature maps before upsampling of models with different attention strategies in Fig.~\ref{fig:ablation}. CW-SA, SW-SA, and CW/SW-SA correspond to the models in the first, second, and third rows of Table~\ref{abl:1}, respectively. We observe that alternately utilizing two self-attention can obtain sharper textures and edges than the other two models. It further demonstrates that the alternate strategy can effectively enhance the expression of features.

\noindent \textbf{Adaptive Interaction Module.} 
We verify the effectiveness of the adaptive interaction module (AIM). 
\textbf{Firstly}, in Table~\ref{abl:2}, we conduct a break-down ablation to investigate the impact of our AIM. The baseline is the model in the third row of Table~\ref{abl:1}, which yields 33.34 dB. Then we introduce a parallel depth-wise convolution (DW-Conv) to self-attention (both SW-SA and CW-SA). The model obtains a 0.07 dB gain over baseline. Finally, we apply the AIM to aggregate two branches and advance the PSNR from 33.41 to 33.52 dB. It proves that our AIM can effectively improve Transformer performance.
\textbf{Secondly}, we further analyze the adaptive interaction between the two branches in Table~\ref{abl:3}. Specifically, our AIM consists of two direction interactions: from SA to Conv (denoted as SA$\rightarrow$Conv), and from Conv to SA (denoted as Conv$\rightarrow$SA). We conduct experiments on three cases: only SA$\rightarrow$Conv, only Conv$\rightarrow$SA, and complete two directions (namely, AIM). The model adopting Conv$\rightarrow$SA outperforms the model using SA$\rightarrow$Conv by 0.04 dB. It means aggregating information to self-attention has a greater impact on performance. And applying the complete AIM gets the best performance. 
These results align with the analysis in Section~\ref{sec:aim}.

\noindent \textbf{Spatial-Gate Feed-Forward Network.}
To illustrate the impact of the spatial-gate feed-forward network (SGFN), we carry out an ablation study in Table~\ref{abl:4}. We compare models using regular FFN~\cite{vaswani2017attention}, SGFN without depth-wise convolution (denoted as SGFN w/o Conv), SGFN without split channel operation (denoted as SGFN w/o Split), and our proposed SGFN. 
\textbf{Firstly}, compared with FFN, utilizing SGFN can effectively reduce the parameters (2.18M) and FLOPs (32.25G) while improving the performance. 
\textbf{Secondly}, the performance is severely degraded when we remove the depth-wise convolution in SGFN. It reveals the significance of spatial information. 
\textbf{Thirdly}, after removing the split operation in SGFN, the PSNR value slightly drops, while the model size and complexity increase a lot. It proves that the information redundancy in channel features impairs the performance of models.

\noindent \textbf{Different Blocks.} 
From the above analyses, we display the effect of each proposed component. We further compare our proposed Transformer blocks, DCTB and DSTB, in Table~\ref{abl:5}. The DCTB and DSTB represent that we replace all Transformer blocks in DAT with DCTB or DSTB. We can discover that the models using single-type blocks have suboptimal performance. The model adopting DSTB performs better than the model using DCTB, aligning with the results presented in Table~\ref{abl:1}. Moreover, both DSTB and DCTB outperform corresponding CW-SA and SW-SA.

\vspace{-1mm}
\subsection{Comparison with State-of-the-Art Methods}
\vspace{-1mm}
We compare our two models, DAT-S and DAT, with the current 11 state-of-the-art image SR methods: EDSR~\cite{lim2017enhanced}, RCAN~\cite{zhang2018image}, SAN~\cite{dai2019secondSAN}, RFANet~\cite{liu2020residual}, HAN~\cite{niu2020singleHAN}, CSNLN~\cite{mei2020imageCSNLN}, NLSA~\cite{mei2021imageNLSA}, ELAN~\cite{zhang2022efficient}, DFSA~\cite{magid2021dynamic}, SwinIR~\cite{liang2021swinir} and CAT-A~\cite{chen2022cross}. Consistent with prior studies~\cite{zhang2018image,liang2021swinir}, we employ a self-ensemble strategy during testing, denoted by the symbol ``+''. Table~\ref{table:psnr_ssim_SR_5sets} presents quantitative comparisons, while Fig.~\ref{fig:sr_vs_2} provides visual comparisons.

\noindent\textbf{Quantitative Results.}
Table~\ref{table:psnr_ssim_SR_5sets} shows results for image SR on factors: $\times$2, $\times$3, and $\times$4. With self-ensemble, our DAT+ outperforms the compared methods on all benchmark datasets with three factors. Meanwhile, DAT performs better than previous methods, except for the PSNR value on the Urban100 dataset ($\times$4) compared with CAT-A. Specifically, compared with SwinIR and CAT-A, our DAT achieves significant gains on the Manga109 dataset ($\times$2), yielding 0.41 dB and 0.23 dB improvements, respectively. Besides, the small vision model, DAT-S, also achieves comparable or better performance compared to previous methods. All these quantitative results indicate that aggregating spatial and channel information from inter-block and intra-block can effectively improve image reconstruction quality. 

\noindent\textbf{Visual Results.}
We show visual comparisons ($\times$4) in Fig.~\ref{fig:sr_vs_2}. In some challenging scenarios, the previous methods may suffer blurring artifacts, distortions, or inaccurate texture restoration. In contradistinction, our method effectively mitigates artifacts, preserving more structures and finer details. For instance, in img\_015, most compared methods hardly recover details and generate undesired artifacts. However, our DAT can restore the correct structures with clear textures. We can find similar observations in img\_047 and img\_049. This is mainly because our method has a more powerful representation ability by extracting complex features from different dimensions.

\begin{table}[t]
\scriptsize
\vspace{1mm}
\begin{center}
\resizebox{\linewidth}{!}{
\setlength{\tabcolsep}{1.2mm}
\begin{tabular}{l|cccccccc}
				\toprule

\rowcolor{color3} & EDSR & RCAN & SwinIR & CAT-A& DAT-S & DAT & DAT-2\\
\rowcolor{color3} \multirow{-2}{*}{Method}& \cite{lim2017enhanced} & \cite{zhang2018image} & \cite{liang2021swinir}  & \cite{chen2022cross} & (ours) & (ours) & (ours)\\
				\midrule
Params (M) & 43.09 & 15.59 & 11.90 & 16.60 & 11.21 & 14.80 & 11.21\\
FLOPs (G) & 823.34 & 261.01 & 215.32 & 360.67 & 203.34 & 275.75  & 216.93\\
Urban100 &26.64 & 26.82 & 27.45 & 27.89 & 27.68 & 27.87  & 27.86 \\
Manga109 &31.02 & 31.22 & 32.03 & 32.39 & 32.33 & 32.51  & 32.41\\
				\bottomrule

\end{tabular}
}
\vspace{-1.5mm}
\caption{\small{Model complexity comparisons ($\times$4). PSNR (dB) on Urban100 and Manga109, FLOPs, and Params are reported.}}
\label{table:complexity}
\end{center}
\vspace{-8mm}
\end{table}

\subsection{Model Size Analyses}
\vspace{-1mm}
We further compare our method with several image SR methods in terms of computational complexity (\eg, FLOPs), parameter numbers, and performance at $\times$4 scale in Table~\ref{table:complexity}. We set the output size as 3$\times$512$\times$512 to compute FLOPs and evaluate performance with PSNR tested on Urban100 and Manga109. Compared with CAT-A~\cite{chen2022cross}, our DAT has comparable or better performance with less computational complexity and model size. Besides, DAT-S obtains excellent performance with lower FLOPs and parameters than SwinIR~\cite{liang2021swinir}. Moreover, to further reveal the better trade-off between model size and performance of our method, we introduce an additional variant model, DAT-2, which is detailed in the supplementary material.

\vspace{-3mm}
\section{Conclusion}
\vspace{-1.5mm}
In this paper, we propose the dual aggregation Transformer (DAT), a new Transformer model for image SR. Our DAT aggregates spatial and channel features in the inter-block and intra-block dual manner, for powerful representation competence. Specifically, successive Transformer blocks alternately apply spatial window and channel-wise self-attention. DAT can model global dependencies through this alternate strategy and achieve inter-block feature aggregation among spatial and channel dimensions. Furthermore, we propose the adaptive interaction module (AIM) and the spatial-gate feed-forward network (SGFN) to enhance each block and realize intra-block feature aggregation between two dimensions. AIM strengthens the modeling ability of two self-attention mechanisms from corresponding dimensions. Meanwhile, SGFN complements the feed-forward network with non-linear spatial information. Extensive experiments indicate that DAT outperforms previous methods.

\noindent \textbf{Acknowledgments}. {This work is supported in part by NSFC grant (62141220, 61972253, U1908212, U19B2035), Shanghai Municipal Science and Technology Major Project (2021SHZDZX0102), and Huawei Technologies Oy (Finland) Project.}

{\small
\balance
\bibliographystyle{ieee_fullname}
\bibliography{egbib}
}

\end{document}